\documentclass[A4paper,12pt]{article}

\pdfoutput=1

\usepackage{latexsym}
\usepackage{epsfig}
\usepackage{verbatim}
\usepackage{multirow}
\usepackage{fancybox}
\usepackage{shadow}
\usepackage{cite}
\usepackage{amssymb}
\usepackage{amsmath}
\usepackage{bm}
\usepackage{amscd}
\usepackage{graphicx}

\usepackage{subfigure}

\author{
}

\title{An optimal learning method for developing personalized treatment regimes}

\vspace{.5cm}

\author{Yingfei Wang \thanks{
       Department of Computer Science, Princeton University, Princeton, NJ  08540, USA, yingfei@cs.princeton.edu }
       \and
	Warren Powell \thanks{
       Department of Operations Research and Financial Engineering, Princeton University, Princeton, NJ  08544, USA, powell@princeton.edu}}

\date{}

\begin{document} \maketitle

\vskip 0.3in

\begin{abstract}

A treatment regime is a function that maps individual patient information to a recommended treatment, hence explicitly incorporating the heterogeneity in need for treatment across individuals. Patient responses are dichotomous  and  can be predicted through an unknown relationship that depends on the patient information and the selected treatment. The goal is to find the treatments  that lead to the best patient responses on average.  
Each experiment is expensive, forcing us to
learn the most from each experiment. We adopt a Bayesian approach  both to incorporate possible prior information and to  update our treatment regime continuously as information accrues, with the potential to allow smaller yet more informative trials and for patients to receive better treatment. By formulating the problem as contextual bandits, we introduce a
knowledge gradient policy  to guide the treatment assignment by maximizing the expected value of information, for which  an approximation
method is used to overcome computational challenges.  We  provide a detailed study on how to make sequential medical decisions under uncertainty to reduce health care costs on a real world knee replacement dataset.   We use clustering and LASSO to deal with the intrinsic sparsity in health datasets. We  show  experimentally that even though the problem is sparse,  through careful selection of physicians (versus picking them at random), we can significantly improve the success rates.

 \end{abstract}

\section{Introduction}
According to the Centers for Medicare and Medicaid Services (CMS),  U.S. health care expenditures grew $5.3$ percent in 2014, reaching $\$3$ trillion or $\$9,523$ per person.  As a share of the nation's Gross Domestic Product, health spending accounted for $17.5$ percent. Rising health care costs have become a major concern for hospital chains \cite{bodenheimer2005high,ginsburg2008high,wennberg2008tracking,lehnert2011review}, which increasingly have to deliver the best care possible within a given budget, forcing them to make better medical decisions.

Common practice is to assign patients to medical professionals (general practitioners, specialists, nurse practitioners) on a first-available basis.  This ignores special expertise with particular medical conditions, as well as the past performance of the physician or facility.  At the same time, physicians may face choices in terms of how to treat a condition, which tends to be guided in part by the past experiences of each physician.  Thus, we have choices of physician (or type of physician), care facility, and specific treatments.  The best choices depend on a combination of the characteristics of the patient, the physician, the facility, and the treatment plan.  We address the problem of how to make the best decisions in the presence of imperfect (and sometimes highly imperfect) understanding of the relationship between patient attributes, medical decisions and medical outcomes.

We are particularly interested in total knee replacement \cite{callahan1994patient, buckwalter1994operative}, a common operation for people with osteoarthritis of the knee which affects more than 27 million people in the U.S. according to the Arthritis Foundation. More than 600,000 knee replacements are performed each year in the U.S., which also leads to extensive post-operative rehabilitation which varies widely from one patient to the next. In order to promote a health care system that provides better care and spends health care dollars more wisely, in 2016, the United States Department of Health and Human Services (HHS) proposed the Comprehensive Care for Joint Replacement (CJR)  where the hospital may be required to repay Medicare for a portion of the cost of a knee replacement episode if the cost and quality fall outside of specified ranges. (HHS 2015) \cite{CMS}.

In this paper, we consider a binary feedback (success/failure) model where if the post-operative cost is below some threshold, the episode of care (spanning initial diagnosis and testing, inpatient treatment and outpatient care) is said to be successful; otherwise it is treated as failure. The aim is to decide the most appropriate physicians and caregivers for each individual patient and maximize the success rate over time.  This is an example of the broader area of personalized medicine,  which formalizes clinical   decision making as a function that maps  individual patient information (including measures of disease stage severity, medical history, clinical diagnosis, genomic information and environmental information) to a recommended treatment. Our work is part of a growing trend toward personalized care where medical decisions are tuned to the characteristics of each patient.  This approach, however, introduces considerable uncertainty in the identification of the best treatments since there is very little data describing patients with the same (or even similar) attributes.  As a result, there is considerable uncertainty in models of the relationship between medical decisions (and patient attributes) with treatment outcomes.  This motivates our work that addresses the problem of balancing between making what appear to be the best decisions, and learning to make better decisions in the future.

There is a variety of new methods to aid in the search for the optimal treatment regime, where a single decision or a series of sequential decisions may be involved, including  sequential multiple assignment randomized trials (SMART) \cite{almirall2012designing, murphy2005experimental},  doubly
robust estimators\cite{murphy2001marginal, zhang2012robust, brinkley2014doubly},  Q-learning \cite{laber2015using, qian2011performance}, adaptive strategies \cite{lavori2000design} and other dynamic treatment regimes studied at length by Robins and colleagues \cite{ murphy2003optimal,  murphy2007customizing,  robins2004optimal,  robins1993information}. Much of the work is trained on past observational data and the treatment regime is not updated with new patient responses.  Yet  when work with historical data, the result is less objective and can be biased substantially due to the differences, for example, in patient populations and in medical facilities or the evolving of the diseases. This is known as offline settings where we are not punished for errors incurred during training and only concern with the final treatment regime after the offline training phases. In this paper, we focus on the case of a single decision and take an online view,   continuously using accumulating
data to modify aspects of
 the treatment regime  as new patients coming in. We adopt a Bayesian approach in which the results from pilot study can be used to construct a reasonable prior treatment regime and which can efficiently  identify any trend,
or optimal clinical benefit as information accures, with the potential  to allow smaller yet more informative trials and for patients to
receive better treatment.

In this regard, it bears similarity with adaptive designs \cite{ashby2006bayesian, berry2010bayesian, jack2012bayesian,chow2014adaptive, bather1985allocation, yin2012phase}. Decisions are made adaptively throughout the running of the trial. Make an observation, update your knowledge, decide what information to collect next is
commonly regarded to be the Scientific Method and the practical guidelines \cite{thall1994practical}. Both our setting and adaptive designs face the same exploration/exploitation dilemma: (1) treat current patients as effectively as possible and (2) have a high probability of correctly identifying the better treatment.      Making what we think is currently the best decision may not be the best given the uncertainty in our model, forcing us to recognize that we have to learn to make better decisions in the future.

In this paper, we formalize personalized medicine as a Bayesian contextual bandit problem. We encounter two challenges.  First, there are very few patients with the same characteristics, which means that it is unlikely that an individual physician sees a sufficient number of eligible patients to produce statistically reliable performance measurements on medical outcomes. We overcome this situation using a parametric belief model that allows us to learn relationships across a wide range of patients and health providers, which is different from  the earlier multi-armed bandit formulations in clinical trials \cite{lai1987adaptive,lai2012efficient} which ignore the attributes of individual patients. Second, due to ethical reasons, testing a treatment decision is expensive.  This puts us in the setting of optimal learning where we need to learn the best treatment as fast as possible. This represents a distinctly different learning environment than what has traditionally been considered using  popular policies such as upper confidence bounding (UCB) which have proven effective in settings with high sampling rates such as learning ad-clicks or the doubly robust estimation which is trained on historical data \cite{dudik2014doubly, zhang2012robust}. We therefore adopt a Bayesian approach and the knowledge gradient policy which takes advantage of domain knowledge to produce rapid learning, and which maximizes success rates for both the current and future patients.

The rest of this paper is organized as follows. In Section \ref{sec:model}, we establish the contextual bandit formularisation for the sequential decision making problems in personalized medicine.  Due to the sequential nature of our problem, we adopt  an online Bayesian logistic regression algorithm in Section \ref{sec:oblr} to handle recursive updates with each patient response. In Section \ref{sec:KG},  we introduce the concept of {\it post-observation} states based on which we develop a knowledge gradient type policy for maximizing the number of successful treatments. In Section \ref{sec:health},  we describe a study investigating the performance of the knowledge gradient policy in the design of health care decisions for knee replacement patients which demonstrates the value of an optimal learning policy to reduce health care costs.   We  conduct feature section through clustering and LASSO to deal with the intrinsic sparsity in the knee replacement data,

\section{Problem Definition}\label{sec:model}
In personalized medicine, a patient arrives characterized by a set of unique characteristics such as measures of disease stage severity, medical history, clinical diagnosis, genomic information, and environmental information, with a health complaint that requires medical intervention.  We use this information as a basis for decisions about medical treatment, including choice of physician, tests, drugs, surgery, and rehabilitation/follow up.  At the end of a treatment episode, we observe a dichotomous health outcome  (e.g. whether cost is below a Medicare-specified threshold, whether the treatment is effective), which is then used to update our understanding of relationships between medical decisions and health outcomes for a patient with a specific set of characteristics.

To design a personalized treatment strategy, the learner is presented with a context vector $\bm{\phi}^X(x^n) =\big(\phi^X_f(x^n) \big)_{f \in \mathcal{F}^X} $ (the characteristics of the $n$th patient $x^n$) and a set of actions $a \in \mathcal{A}$ (doctors, treatments, rehabilitation).  Each action  $a$ is associated with a feature vector $\bm{\phi}^A(a)$.   Discrete treatments are handled with indicator variables (such as $ \mathbb{I}(a,``drug'') = 1$ if an attribute $a$ refers to administering a particular drug).  After choosing an action $a$, a response of patient $y^{n+1}\in \{-1, +1\}/$  \{failure, success\} for the action $a$ is revealed, but the rewards of other actions are not observable.   The ``success'' or ``failure'' may 
depend stochastically on $x$ and $a$.

A policy $\pi$ or a treatment regime is a function mapping from any context information $x$ to an action $a$. We denote the `patient horizon' as $N$ which is the number of present and future patients who will be treated with one of the treatment in $\mathcal{A}$. It is worth noting that $N$ need not to be known beforehand  and may depend on the pattern of a emerging disease, the performance of current treatments, the emergence of new treatments,  and may be infinite for recurring conditions such as knee replacements.

We adopt probabilistic modeling for the unknown probability of success. Under general assumptions, the posterior probability of class $+1$ can be written as a  link function acting on a linear function of the feature vector 
$$
p(y=+1|x, a)=\sigma \Big( w_0 + \sum_{f \in \mathcal{F}^X}w_f^X\phi_f^X(x) +\sum_{f \in \mathcal{F}^A}w_f^A\phi_f^A(a)\Big),
$$ 
with the link function $\sigma(b)$ often chosen as the logistic function $\sigma(b)=\frac{1}{1+\text{exp}(-b)}$ or probit function $\sigma(b)=\Phi(b)=\int_{-\infty}^b\mathcal{N}(s|0,1^2) \text{d}s.$    The main difference between the two sigmoid functions is that the logistic function has slightly heavier tails than the normal CDF.  We can make this more compact by introducing the basis function $\phi_0 = 1$. Now let $\bm{\phi}(x,a)$ be a column vector of basis functions, and $\bm{w}$ be a column vector of the coefficients. The probability of class $+1$ can be re-written as 
$$p(y=+1|x,a) = \sigma(\bm{w}^T\bm{\phi}).$$

We adopt a Bayesian view and start with a multivariate prior distribution for the unknown parameter vector $\bm{w}$.  We use $S^n = (K^n, x^n)$  to  denote  the state of the system at time $n$ which includes the ``state of knowledge'' $K^n$  that captures our belief about the parameters and the context information $x^n$. Each of the past observations are made of triplets $(x^n, a^n,y^{n+1})$, assuming labels $y$ are generated independently.  Let  $\mathcal{D}^n=\{(x^i, a^i,y^{i+1})\}_{i=0}^{n-1}$  denote the previous measured data set  for any $n=0,\dots,N-1$.   Note that the notation here is slightly different from the (passive) PAC learning model  where the data  are i.i.d. and are denoted as $\{(x_i, y_i)\}$. Yet in our (adaptive) sequential decision setting, measurement  $a^{n}$ depends on the state $S^n$, while $Y^{n+1}$ is a random variable that has not been observed at time $n$. This notation with superscript indexing time stamp is standard, for example,  in control theory, stochastic optimization and optimal learning. A history of the process can be represented using 
$$
h^n = (K^0, x^0, S^0, a^0, Y^1, K^1, x^1,S^1,a^1, Y^2,...,a^{n-1}, Y^n,K^n, x^n, S^n).
$$
We  use Bayes' theorem to form a sequence of posterior predictive distributions $  p(\bm{w}|\mathcal{D}^n)$ for $\bm{w}$ from the prior and the previous measurements.

 The  goal is to find a policy that selects actions such that the cumulative reward is as large as possible over time, or equivalently, treatment on patients is as effectively as possible:
 
\begin{equation}\label{obj}
\max_{\pi}\mathbb{E}\bigg[\sum_{n=0}^{N-1}Y^{n+1}\Big(S^n, {A^{\pi}(S^n)}\Big)|S^0\bigg],
\end{equation}
where $Y$ denotes the random variable of the patient response, $A^{\pi}$ denotes the treatment recommended by the dynamic treatment regime $\pi$.

\section{Online Bayesian logistic regression based on Laplace approximation} \label{sec:oblr}
A Bayesian approach to linear classification models  requires a prior distribution for the weight parameters $\bm{w}$, and the ability to compute  the conditional posterior $p(\bm{w}| \mathcal{D})$ given the observation. Specifically, suppose we begin with an arbitrary prior $p(\bm{w})$
and apply Bayes' theorem to calculate the posterior:
$$
p(\bm{w}|\mathcal{D}) =\frac{1}{Z} p(\mathcal{D}|\bm{w})p(\bm{w}),
$$ where the normalization constant $Z$ is the unknown evidence. An $l_2$-regularized logistic regression can be interpreted as a Bayesian model with a Gaussian prior on the weights with standard deviation $1/\sqrt{\lambda}$.

Unfortunately, exact Bayesian inference for linear classifiers are intractable since the evaluation of the posterior distribution comprises a product of sigmoid functions; in addition,  the integral in the normalization constant is intractable as well. We can either use analytic approximations to the posterior, or solutions based on Monte Carlo sampling, foregoing a closed-form expression for the posterior. Observations come one by one due to the sequential nature of our problem setting. After each new observation,  retraining the Bayesian classifier using all the previous data is computationally inefficient in terms of both time and space complexity. To this end, we use the online Bayesian linear classification algorithm proposed by Wang {\it et al.} \cite{wangKG2016} handle recursive updates with each patient response.

Due to its equivalence to $l_2$ regularization, independent normal priors (with $ \bm{\Sigma}=\lambda^{-1} \bm{I}$, where $ \bm{I}$ is the identity matrix) are considered. Laplace approximation to the posterior is used to make the computation tractable. In our model, the posterior is approximated by a Gaussian approximation with diagonal covariance matrices. It can be obtained by finding the mode of the posterior distribution and then fitting a Gaussian distribution centered at that mode. Specifically, define the logarithm of the unnormalized posterior distribution 
 \begin{eqnarray}\nonumber 
 \Psi(\bm{w})=\log p(\mathcal{D}|\bm{w})+
\log p(\bm{w}). \end{eqnarray}
 The Laplace approximation is based on a second-order Taylor expansion to $\Psi$ around its MAP (maximum a posteriori) solution $\hat{\bm{w}}= \arg \max_{\bm{w}}\Psi(\bm{w})$: 
 \begin{equation}\label{exp}
 \Psi(\bm{w}) \approx \Psi(\hat{\bm{w}})-\frac{1}{2}(\bm{w}-\hat{\bm{w}})^T \bm{H}(\bm{w}-\hat{\bm{w}}),
 \end{equation}
 where $\bm{H}$ is the Hessian of the negative log posterior evaluated at $\hat{\bm{w}}$:
 $$\bm{H}=-\nabla^2 \Psi(\bm{w})|_{\bm{w}=\hat{\bm{w}}}.$$
 
 By exponentiating both sides of Eq. \eqref{exp}, we can see that the Laplace approximation results in a normal approximation to the posterior 
\begin{equation*}\label{pos}
p(\bm{w}|\mathcal{D}) \approx \mathcal{N}(\hat{\bm{w}},\bm{H}^{-1}).
\end{equation*}

In our setting, the use
of this convenient approximation of the posterior is twofold. It first serves as a prior on the weights
to update the model when a new patient response becomes available. Second, it defines the belief states in the Bayesian policies, for example, the knowledge gradient policy and Thompson sampling that we introduce later.  Starting from Gaussian priors $\mathcal{N}\big(w_j|m_j^0, (q_j^0)^{-1}\big)$ over $w_j$ with mean $m_j^0$ and variance $(q_j^0)^{-1}$, after the first $n$ patient responses, the Laplace approximated posterior distribution is $p(w_j|\mathcal{D}^n) \approx \mathcal{N}\big(w_j|m^n, (q_j^n)^{-1}\big)$. At the $n$th time step, we  find the MAP  solution \eqref{exp} to the posterior after the new information $(x^n, a^n, y^{n+1})$ by the one-dimensional bisection method \cite{wangKG2016}:
 \begin{equation}\label{m}
  \bm{m}^{n+1} = \arg \max_{\bm{w}}\frac{1}{2}\sum_{j=1}^d q_i^n(w_i-m_i^n)^2 +\log\Big( 1+ \exp\big(-y^{n+1}\bm{w}^T\bm{\phi}^n\big)\Big),
  \end{equation}
  where $\bm{\phi}^n$ is a compact notation for $\bm{\phi}(x^n, a^n)$.
 The inverse variance of each weight $w_j$ is given by the curvature at the mode as:
 \begin{equation}\label{q}
 q_j^{n+1} = q_j^n +\zeta^n(1-\zeta^n)( \bm{\phi}^n_j)^2,
 \end{equation}
 where $\zeta^{n} = \Bigg(1+\exp \Big ((\bm{w}^{n+1})^T \bm{\phi}^n \Big)\Bigg)^{-1}$.

\section{Knowledge gradient policy}\label{sec:KG}
The knowledge gradient (KG) policy, first proposed for offline ranking and selection problems, maximizes the value of information from a decision.
In ranking and section problems,  the performance of each alternative is represented by a (non-parametric) lookup table model. Although originally developed for offline learning (where we do not pay attention to successes while we are learning), it is easily adapted to online learning where we seek to maximize the cumulative number of successes. After its first appearance, KG has been extended to various belief models (e.g.  \cite{mes2011hierarchical,negoescu2011knowledge,ryzhov2012knowledge,wang2015nested, wangKG2016}). 
 Experimental studies have shown good performance of the knowledge gradient policy in  learning settings with expensive experiments, especially in early iterations \cite{wang2015nested, wangKG2016, russo2014learning, ryzhov2012knowledge}. This is particularly well suited to personalized medicine where we want to learn as fast as possible from each patient response so as to provide better treatment on the upcoming patients. In comparison, other policies like upper confidence bounding (UCB) is known to explore more than necessary, leading to unnecessarily poor treatment performance on current patient. 

\subsection{Markov decision process formulation}
We define the knowledge state in our setting as $K^n = (\bm{m}^n, \bm{q}^n)$. In a dynamic program, the value function is defined as the value of the optimal policy given a particular state $S^n \in \mathcal{S}$ at time $n$, and may also be determined recursively through Bellman's equation.   At time $N$, we should simply choose the alternative
that looks the best given everything we have learned, because there are no longer any future
decisions that might benefit from learning. Since the goal of personalized medicine is to maximize the cumulative success of treatment, the terminal value function $V^N: \mathcal{S} \mapsto \mathbb{R}$ is given by
$$V^N(s) = \max_{a}\mathbb{E}_Y\bigg[Y^{N+1}(s,a)\big|s\bigg] = \max_a p(y=+1|s,a), \forall s \in \mathcal{S}.$$ 

The value function at any other time $n=0,1,...,N-1$ is given recursively by Bellman's equation for dynamic programming:
\begin{equation}\label{bellman}
V^n(S^n) = \max_{a} \mathbb{E}_{x,Y}\bigg[Y^{n+1}(S^n,a)+V^{n+1}(S^{n+1})|S^n\bigg].
\end{equation}

We write Bellman's equation in its standard form (Eq. \eqref{bellman}) by writing the sequence of states, actions and information using:
$$(S^0,a^0,Y^1,S^1,a^1,Y^2, \ldots, S^n,a^n,Y^{n+1}, \ldots )$$
When we include contextual information, this sequence would be written
$$\Big(K^0,x^0,S^0=(K^0,x^0),a^0,Y^1,K^1,x^1, S^1=(K^1,x^1),a^1,Y^2, \ldots \Big)$$
Here, the knowledge state $K^n$ might be called the {\it post-observation} state, while $S^n$ is the {\it pre-decision} state, representing the state after a patient has arrived.  Instead of relating $V^n(S^n)$ to $V^{n+1}(S^{n+1})$ as is classically done in Bellman's equation, we can break the recursion into two steps: from $S^n$ to $K^{n+1}$, and then from $K^{n+1}$ to $S^{n+1}$, giving us the following two-step version of Bellman's equation.
\begin{eqnarray}
V^n(S^n) &=& \max_a \mathbb{E}\Bigg\{Y^{n+1}(S^n, a) + V^{k,n+1}\bigg(K^{n+1}\Big(Y^{n+1}(S^n, a)\Big)\bigg) \big|S^n\Bigg\},\\\label{x}
V^{k,n+1}(K^{n+1})&=& \mathbb{E}_x\Bigg[V^{n+1}(S^{n+1})| K^{n+1}\Bigg ].
\end{eqnarray}

Bellman's equation works well for problems with small state and action spaces, and where
the transition matrix can be easily computed.  But in personalized health care, the context information $x^n$ can only be observed (the distribution is unknown). The context information can be an arbitrary sequence which is fixed beforehand or stored in historical data, or it can be  non-stochastically chosen by an adversary. The attractiveness of the post-observation state that the maximum and the expectation over context $x$
are interchanged, giving us computational advantages to use  simulation based approach without    probabilistic modeling  of the contextual information and by treating the contextual information as arbitrarily  given by the oracle.

\subsection{Knowledge gradient policy with contextual information}
In order to approximately solve the Bellman's equations in the previous section, we develop the knowledge gradient policy around the values in {\it post-observation} states $K^n$. The knowledge gradient  $\nu_{a}^{\text{KG},n}(S^n)$ of measuring an action $a$  in state $S^n=(K^n, x^n)$ is defined as the single-step expected improvement in value if action $a$ is taken. 
\begin{equation} \label{KG}
\nu_{a}^{\text{KG},n}(S^n) := \mathbb{E}_Y\Bigg[ V^{k,N}\Big(K^{n+1}\big(Y^{n+1}(S^n,a)\big)\Big)-V^{k,N}(K^n)|S^n\Bigg],
\end{equation}
where $K^{n+1}\big(Y^{n+1}(S^n,a)\big)$ is the next stochastic state of knowledge if we choose treatment $a^n =a$ right now, allowing us to observe the stochastic patient response $Y^{n+1}$. This allows us to update $\bm{m}^n$ and $\bm{q}^n$ based on Eq. \eqref{m} and Eq. \eqref{q}, transitioning to the next state of knowledge $K^{n+1}$. The knowledge gradient policy then balances the treatment that appears to be the current best and the one that learns the most by choosing an action $a$ at time $n$ as \cite{wangKG2016}:
$$A^{\text{KG},n}(S^n) = \arg \max_{a} p(y=+1| S^n,a) + \tau \nu_{a}^{\text{KG},n}(S^n),$$ where $\tau$ reflects a planning horizon that captures the value of the information we have gained on future patients.

Given a knowledge state $k=(\bm{m},\bm{q})$, or equivalently $p(w_j|k)=  \mathcal{N}(w_j|m_j, q_j^{-1})$, the predictive Bernoulli distribution of patient response $y$ for a context $x$ and a treatment $a$ can be found by marginalization over $\bm{w}$,
\begin{eqnarray}\label{predictD}
 p(y=+1|x,a,k) =\int  p(y=+1|x,a, \bm{w}) p(\bm{w}|k)\text{d}\bm{w}
= \int \sigma\big(\bm{w}^T\bm{\phi}(x,a)\big)p(\bm{w}|k)\text{d}\bm{w}.
\end{eqnarray}  
For notational simplicity, we drop $\bm{\phi}$'s dependence on $x$ and $a$. Denoting $b = \bm{w}^T\bm{\phi}$ and $\delta(\cdot)$ the Dirac delta function, we have  $\sigma(\bm{w}^T\bm{\phi}) = \int \delta(b - \bm{w}^T\bm{\phi}) \sigma(b) \text{d}b$. Hence we have 
 $$\int \sigma(\bm{w}^T\bm{\phi})p(\bm{w}|k)\text{d}\bm{w} = \int \sigma(b) p(b) \text{d}b,$$
 where $p(b) = \int \delta(b - \bm{w}^T\bm{\phi})p(\bm{w}|k) \text{d}\bm{w}.$ Since the delta function imposes a linear constraint on $\bm{w}$ and  $p(\bm{w}|k)$ is Gaussian, the marginal distribution $p(b)$ is also Gaussian. We can evaluate $p(b)$ by calculating the mean and variance of this distribution. We have 
 \begin{eqnarray*}
 \mu_b&=&\mathbb{E}[b]=\int p(b)b \text{ d}b = \int p(\bm{w}|k)\bm{w}^T\bm{\phi} \text{ d}\bm{w}=\bm{m}^T\bm{\phi},\\
\sigma_b^2&=& \text{Var}[b]=\int p(\bm{w}|k) \big((\bm{w}^T\bm{\phi})^2-(\bm{m}^T\bm{\phi})^2 \big) \text{ d}\bm{w}= \sum_{j=1}^d q_j^{-1} \phi_j^2.
 \end{eqnarray*}
 Thus $\int \sigma\Big(\bm{w}^T \bm{\phi}(x,a)\Big)p(\bm{w}|k)\text{d}\bm{w} = \int \sigma(b) p(b) \text{d}b = \int \sigma(b)\mathcal{N}(b|\mu_b, \sigma_b^2).$
 Since the convolution of a Gaussian $\mathcal{N}(b|\mu_b, \sigma_b^2)$ with a logistic function $\sigma(b)$ cannot be evaluated analytically, we apply the approximation $\sigma(b) \approx \Phi(\alpha b)$ with $\alpha = \pi/8$. Denoting $\kappa(\sigma^2)=(1+\pi \sigma^2/8)^{-1/2}$ ,  we have
 $$
 p(y=+1|x,a, k)=\int \sigma \Big(\bm{w}^T\bm{\phi}(x,a)\Big)p(\bm{w}|k) \text{d}\bm{w} \approx \sigma(\kappa(\sigma^2_b)\mu_b).
$$

We are now ready to compute the knowledge gradient value  $\nu_{a}^{\text{KG},n}(S^n)$ in Eq. \eqref{KG}. First, $V^{k,N}(K^n)$ is deterministic at time $n$. Since the patient response $Y^{n+1}$ is not known at the time of selection, the expectation is computed over the Bernoulli distribution and the current belief model specified by $K^n$.  Specifically, the expectation can be obtained by averaging the two possible responses $+1/-1$ as follows:
\begin{eqnarray*}
&&\mathbb{E}\Big[V^{k,N}\big(K^{n+1}(Y^{n+1}(S^n,a))\big)|S^n\Big] \\&=&  p(y=+1|x,a, K^n)V^{k,N}\Big(K^{n+1}_{+1}(S^n,a)\Big)+  p(y=-1|x,a, K^n)V^{k,N}\Big(K^{n+1}_{-1}(S^n,a)\Big)\\
&=& p(y=+1|x,a, K^n)\cdot  \max_{a'} p\Big(y = +1 | x,a',K^{n+1}_{+1}(S^n,a)\Big) \\
&&+ p(y=-1|x,a, K^n)\cdot  \max_{a'} p\Big(y = +1 | x,a',K^{n+1}_{-1}(S^n,a)\Big),
\end{eqnarray*}
where  $K^{n+1}_Y(x,a)$ denotes the next belief state given outcome Y according to Eq. \eqref{m} and Eq. \eqref{q}.

It can be seen that the Laplace approximation and the recursive update make the computation of the knowledge gradient tractable by analytically approximating the value in the next state $V^{n+1}(S^{n+1})$ and offering computational simplicity with Gaussian distributions.

\section{A case study on the knee replacement  } \label{sec:health}
\begin{figure}[htp!]
\centering
\includegraphics[width=0.5\textwidth]{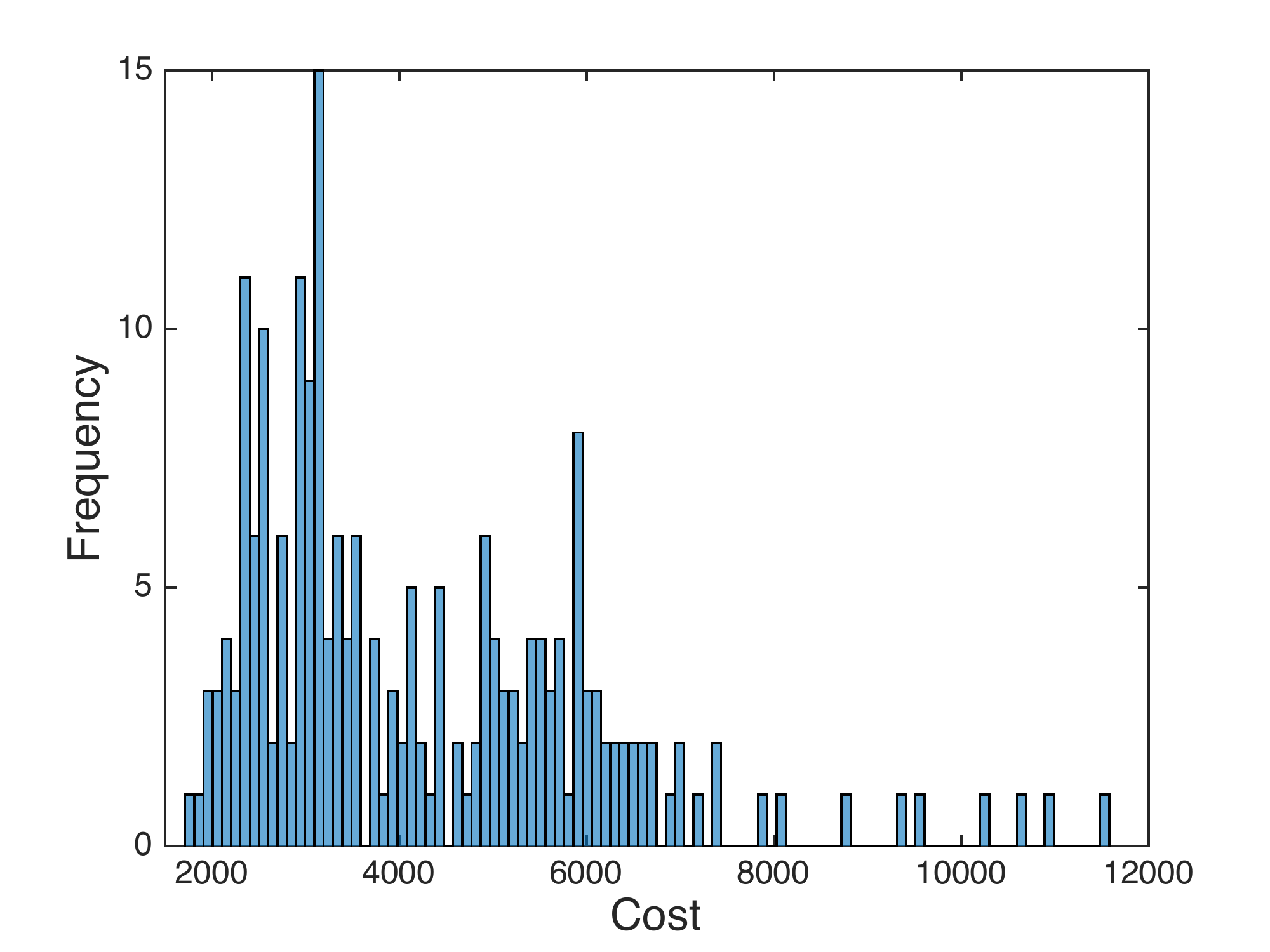}
\caption{Post-operative cost distribution. \label{Abs} \label{cost}}
\end{figure}

In the knee replacement patient datasets, we have 212 episodes involving 26735 records of patient information with either icd9 diagnosis, icd9 procedure or hcpcs/cpt procedure record type, where icd9 is the ninth revision of the International Classification of Diseases and the Healthcare Common Procedure Coding System (hcpcs) is a set of health care procedure codes based on the American Medical Association's Current Procedural Terminology (cpt).  To make a fair statement of the costs, all the 212 episodes are obtained from the same health care provider. The episode ID, beneficiary ID and claim ID have been replaced with randomly assigned numbers for anonymization.   There are procedures that occur before the knee replacement, and procedures after the replacement that constitute rehabilitation. We see a variety of different post-operative costs of 212 episodes, ranging from $\$1787.69$ to $\$11571.44$, as depicted in Fig. \ref{cost}. We adopt  our success/failure model that if the cost is smaller than the 2/3 quantile of the costs ($\$ 4937.23$), we think it as acceptable, otherwise we treat it as failure.  The challenge is how the physicians and caregivers affect the post-operative costs.  

A patient can have a number of attributes, spanning personal characteristics (age, weight, gender, ethnicity, body type), their medical history, and diagnostic information.  We first translate the data file into a flat file. Specifically,  we create a set of columns consisting of every  diagnosis (icd9 diagnosis code) and/or pre-operative procedure (icd9 procedure or hcpcs/cpt procedure code) that has appeared in the dataset. For each patient,   mark the value of that column as 1 if the patient has a diagnosis or uses the procedure,  otherwise mark as 0.   Together with the demographic characteristics  (sex and age), these constitute the feature vector of a patient and thus the context to the contextual bandit problems. 

Not surprisingly, there are a huge number of features associated with each patient, e.g. 979 columns of possible diagnoses and 772 columns of pre-operative procedures. These features are typically sparse -- compared to nearly 2000 features, the number of 1's for each patient ranges from 8 to 110, with an average of 31.   For this reason, we use $l_2$ regularization in our Bayesian logistic regression to handle the sparsity of the model.  One way of learning the sequential physicians/caregivers assignment is by separating the 212 episodes into two sets. We can use one set to generate a prior distribution on the weight vector $\bm{w}$ and use the other set for online learning of  physicians/caregivers assignments. Yet if we directly use  these features, the  sparsity and the relative small number of patients makes learning more difficult and is computationally expensive.  Besides, simplification of models can make them easier to interpret by researchers and enhance generalization by reducing overfitting. We instead find the lower dimension feature representation as explained in the next section.

\subsection{Feature selection}
There are several ways to look for more compact representations, such as principal component analysis (PCA), singular value decomposition (SVD) or auto encoder to perform the dimension reduction. However, the features from the non-linear dimension reduction lose the original meanings of the health terminologies.  Yet in health care analytics, interpretability of the resulting feature subspace is  desired. For example, it is interesting to know whether age or malignant essential hypertension affect the cost or quality of total knee replacement.

\begin{figure}[htp!]
\centering
\includegraphics[width=0.63\textwidth]{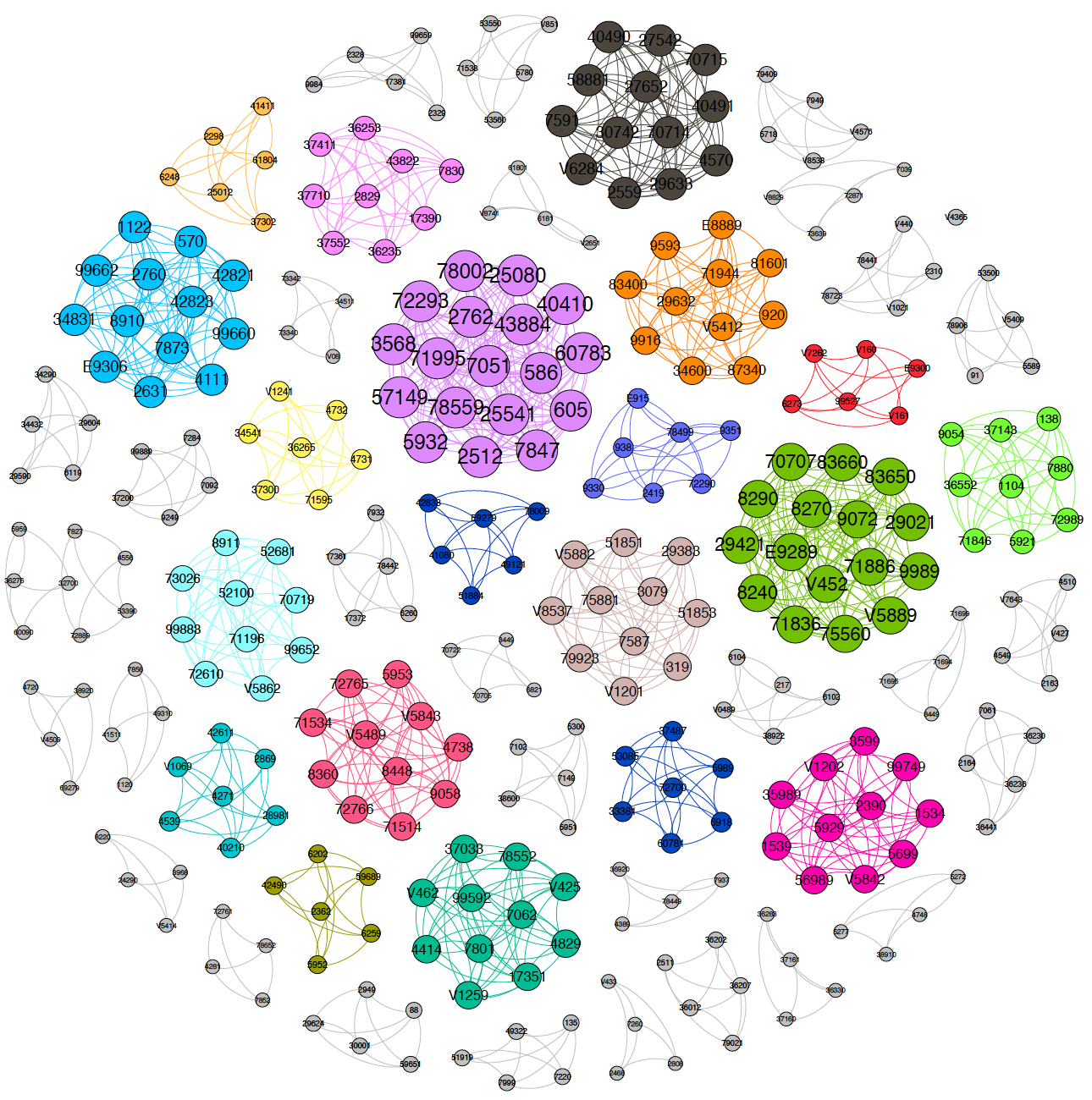}
\caption{Cluster of the diagnoses.  \label{network}}
\end{figure}

Due to the nature of health analytics, many features  tend to happen at the same time.  For example, in terms of diagnoses, obesity and hypertension are linked, with obese patients having higher rates of hypertension than normal-weight individuals \cite{chiang1969overweight}. In addition, certain tests are often run together. In order to capture this characteristic, we first  cluster the diagnoses into groups based on their occurrences.  We construct an undirected network to represent the relationship of different diagnoses as follows.  We treat each icd9 diagnosis code as a node in the network. We measure the occurrence similarity of any diagnosis pair $(d_1,d_2)$ by their intersection angle.  Specifically, each diagnosis is represented by a 212 dimensional binary vector indicating whether  each patient has that diagnosis.  We then set a threshold of the cosine of the intersection angle $\frac{\langle\,d_1, d_2\rangle}{\|d_1\|\|d_2\|}$.  For a diagnosis pair $(d_1,d_2)$, if the cosine of the intersection is larger than the threshold,  we draw  an edge between them. It is worth noting that if we set the threshold to be 1, it is equivalent to saying that $d_1$ and $d_2$ always  happen at the same time across all the patients if there is an edge between them.  When we set the threshold to $.8$,  the presence of an edge means that $d_1$ and $d_2$ are recorded together 80 percent of the time. 

After the construction of the network, we find the clusters/groups by detecting the weakly connected components in the network. In Fig. \ref{network}, each node is labeled with its icd9 diagnosis code with the size of the node corresponding to its degree. Different colors represent different groups/cliques. The nodes with degree less than 3 are filtered. After clustering, 979 diagnoses have been grouped into 608 components.  For example, the red group on the upper right consists of icd9 diagnosis code V160 (Family history of malignant neoplasm of gastrointestinal tract), V161 (Family history of malignant neoplasm of trachea, bronchus, and lung), 99527 (Other drug allergy),  6273 (Postmenopausal atrophic vaginitis),   E9300 (Penicillins causing adverse effects in therapeutic use), V7262 (Laboratory examination ordered as part of a routine general medical examination).

\begin{figure}[htp!]
\centering
\includegraphics[width=0.5\textwidth]{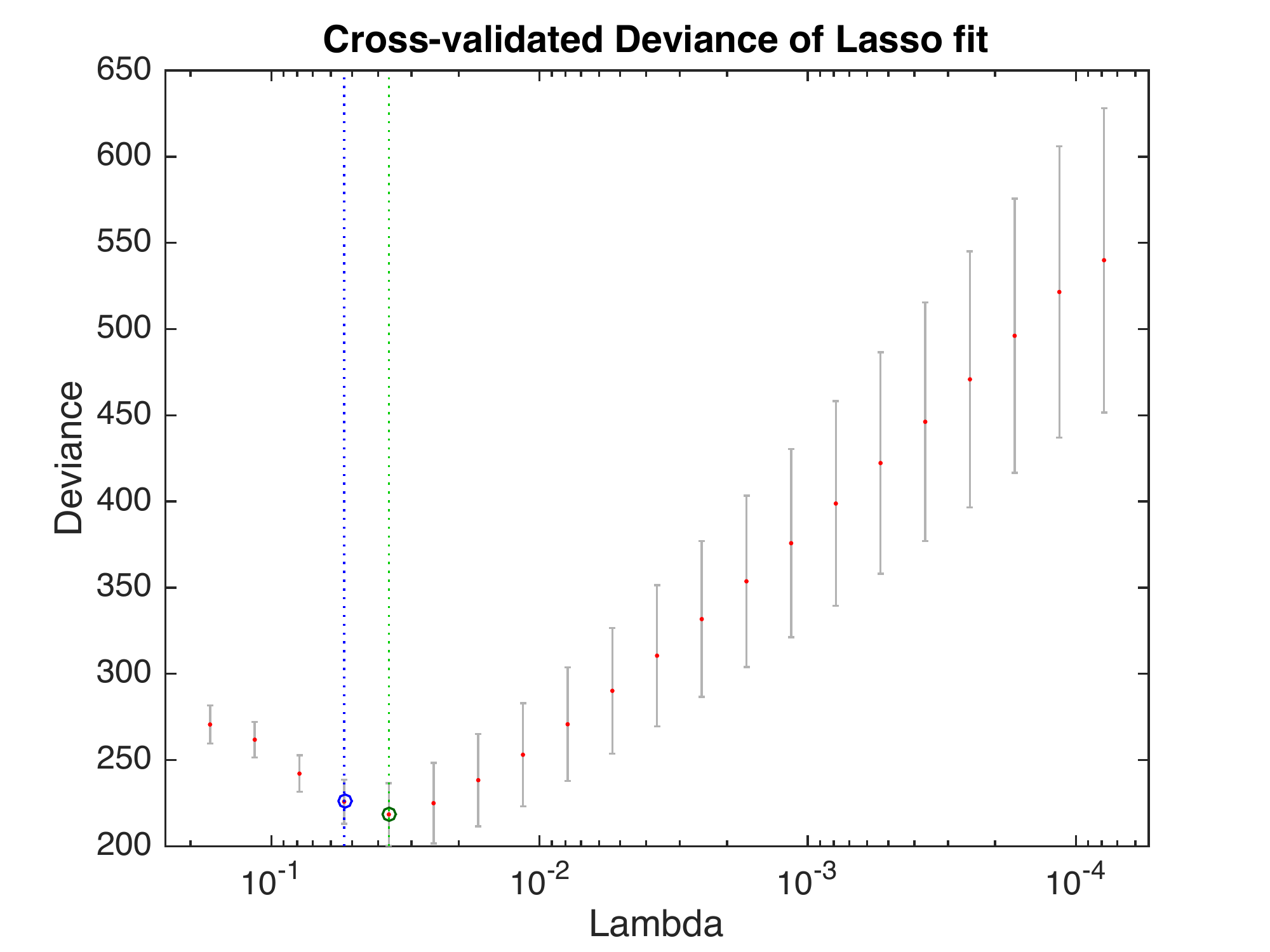}
\caption{Cross-validated deviance plot.  \label{val}}
\end{figure}

Instead of using individual diagnosis codes as the features for the patients, we first use the clusters as the features.  We further conduct feature selection by selecting a subset of relevant features for use in model construction.  Specifically, we use the $l_1$  regularized (LASSO) logistic regression to yield the sparse solutions  of selecting a subset of relevant features using 25  regularization parameter  (Lambda) values and 10-fold cross validation on the patient datasets.   Fig. \ref{val} identifies the minimum-deviance point with a green circle and dashed line as a function of the regularization parameter. The blue circled point has minimum deviance plus no more than one standard deviation. We use the 31 selected features at the blue line as the set of relevant patient attributes to proceed our contextual bandit learning in the next section.  Many other interesting statistical questions can be asked regarding this dataset, e.g. feature importance, best prediction model or statistical significance. Yet they are not the main focus of this work which addressed the optimal learning challenge  with stochastic binary feedback.

\subsection{Personalized physicians and caregivers assignment}

We discovered from data that not surprisingly  there is a number of caregivers (with national provider identifier NPIs) performing the rehabilitation.  The caregivers should be divided naturally into communities or modules. Different people doing rehab on the same patient will belong to the same facility. Since some   facilities keep patients longer than other ones,   in this case, what is important is the facility, not the individual caregiver.  If we can detect and characterize this community structure,  this should give us a set of "facilities" (in the form of these clusters).  We  use the same idea as in the previous section to construct a network that represents the relationships between different caregivers. Each node is one caregiver. Since two caregivers from the same  facility do not necessarily always come to the same patients, we lower the threshold of the  cosine of their intersection angle to  0.5  to capture this reality. 
\begin{figure}[htp!]
\centering
\includegraphics[width=0.9\textwidth]{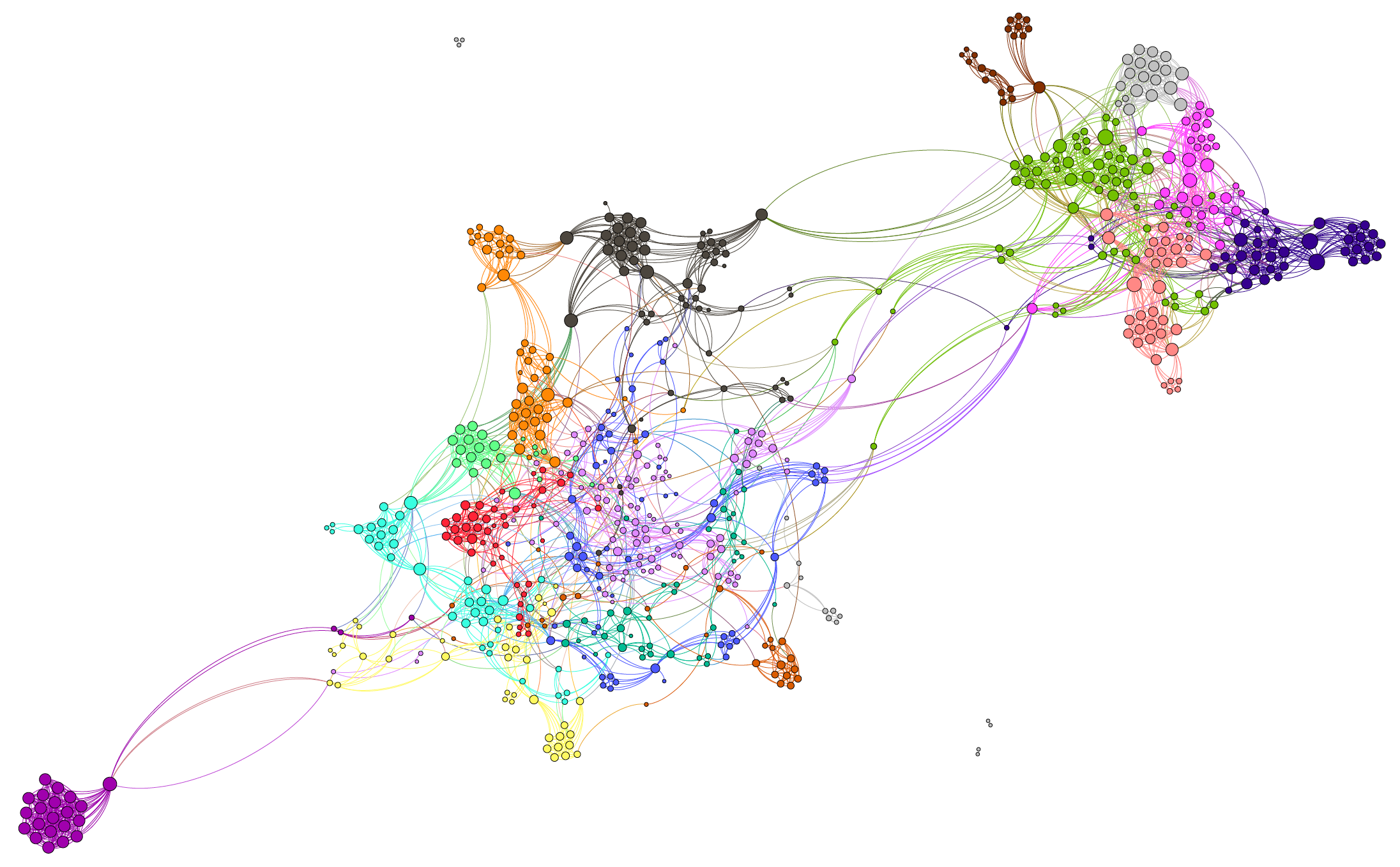}
\caption{Clustering of the caregivers.  \label{module}}
\end{figure}

 In the previous section, when the threshold is 1, the edge represents an equivalence relation (reflexive, symmetric and transitive), making each weakly connected components  a clique. Yet when lowering the threshold, it is less meaningful to treat weakly connected components  as a group. We instead detect the modularity of the total 768 caregivers using spectral community detection \cite{newman2006modularity}. Fig. \ref{module} depicts the modular classes (58 in total) and different classes are drawn with different colors.  Network visualization provides strong hints of connectional relationships. Nodes within each modular class have dense connections with each other while inter-modular-class connections are sparse. 

We now study the problems of how physicians and facilities (rather than each caregiver) affect the cost of each patient.  We represent each physician/facility by an indicator variable and the Bayesian linear classification belief models can be thus presented as $\sigma \Big(F \big(x,(p,f) \big) \Big)$ with $p$ and $f$ denote the chosen physician and facility, respectively,
\begin{equation}\label{pf}
 F \big(x,(p,f) \big) = w_0+ (\bm{w}^X)^T \bm{\phi}^X(x) + \sum_{m=1}^M w^P_m \mathbb{I}(p,p_m) +  \sum_{l=1}^Lw^F_l\mathbb{I}(f,f_l).\end{equation}
 For each patient, one and only one $p$ (or $f$) can be assigned such that  exactly  one $\mathbb{I}(p,p_m)$ is 1.

We sort the patient data chronologically. For each patient visit, based on the patient attributes $x$, we assigned a physician for the surgery and/or a facility for the rehab. We then receive a payoff of whether it is success or failure.  The goal is to maximize the number of successes across all 211 patients. There is a fundamental exploration vs.
exploitation tradeoff: in order to learn the success rate of each physician/facility, it needs to be tried for long time benefit, leading to a potential drop in the short-term performance. 

Evaluating an exploration/exploitation policy is difficult
since we do not know the outcome for physicians and facilities that were not chosen for a particular patient in the record data. Based on the real world context and patient features in
the knee replacement dataset, we instead simulate the true outcomes using a weight vector $\bm{w}$.   We use our ability to simulate the true $w$ to compute the true probability of a successful outcome, and compare this to our simulated success rate based on decisions from our policy using the estimated model.  Although we have modeled the choice of physician and rehabilitation facilities, these are handled in an identical way as in Eq. \eqref{pf}, and as a result we focused just on the choice of physician.  

We set the number of available physicians as $M=20$. The experimental results are reported on 500 repetitions of each algorithm. The only difference is the way each policy selects the actions; all the
rest, including the model updates, is identical as described
in Section \ref{sec:oblr}.

Binary feedbacks (such as success/failure outcomes) are inherently noisy, which creates a problem for value-of-information policies such as the knowledge gradient.  The problem is that we learn very little from a single outcome.  Now consider what happens if we decide to test, say, $k$ patients.  The value of information from $k=1$ patients may be quite low, but the value can grow nonlinearly (specifically, in the shape of an S-curve), producing much greater value if we are willing to consider the combined information learned from, say, $k=20$ patients \cite{Frazier:2010:PLM:1898671.1898677}.  As a result of this non-concavity in the value of information, we propose to consider the impact of  \textbf{posterior reshaping} for the KG policy. In particular, for normally distributed posteriors on $\bm{w}$, decreasing the variance would have the effect of increasing exploitation over exploration. In our simulations, we have tried to reshape the covariance matrix $\bm{\Sigma}^n$ to $\eta^2\bm{\Sigma}^n$. This only affects the calculation of the knowledge gradient and does not change the model updates. 
 
 Posterior reshaping has an effect similar to the  KG$(*)$ policy. The KG$(*)$ policy mimics the effect of doing a single batch of $n_x$ samples, which is the same as replacing an experiment with precision $\beta^W$ with one where the precision is $n_x \beta^W$ (higher precision means lower uncertainty).  The idea behind the KG$(*)$ policy is to find $n_x$ that produces the highest {\it average} value of information, which  effectively finds  the tangent point of the value of information of measuring a single alternative $x$ for many times \cite{Frazier:2010:PLM:1898671.1898677}. Yet finding the  tangent point is a heuristic and the KG$(*)$ policy implicitly assumes that our budget is large enough to sample alternative $x$ roughly $n_x^*$ times. In comparison, posterior reshaping can also be understood as (hypothetically) repeated experiments for each alternative and the tunable parameter $\eta$ offers great computational efficiency and flexibility (e.g. it is not restricted to the   tangent point).
 
 We compare our policy with pure exploitation (which assigns the physician that seems to be the best), pure exploration (which randomly assign a physician, as would happen if you assigned the first available physician) and Thompson sampling \cite{thompson1933likelihood}.  Thompson sampling has been successfully applied to two-treatment adaptive designs \cite{berry1995adaptive, hu2006theory} and other applications \cite{graepel2010web, agrawal2012thompson}. In our personalized health settings, at each time step $n$, given the patient information $x^n$, it first draws a sample $\hat{\bm{w}}$ according to the posterior distribution $p(\bm{w}|\mathcal{D}^n)$. It then selects a treatment $a^n$ (that is, the physician) that maximizes the probability of success under the sample parameter value $a^n = \arg\max_{a}p(y=+1|x^n,a,\hat{\bm{w}})$.
 
 \begin{figure}[htp!]
\centering
  \subfigure[Cumulative successes. \label{patient}]{\includegraphics[width=0.45\textwidth]{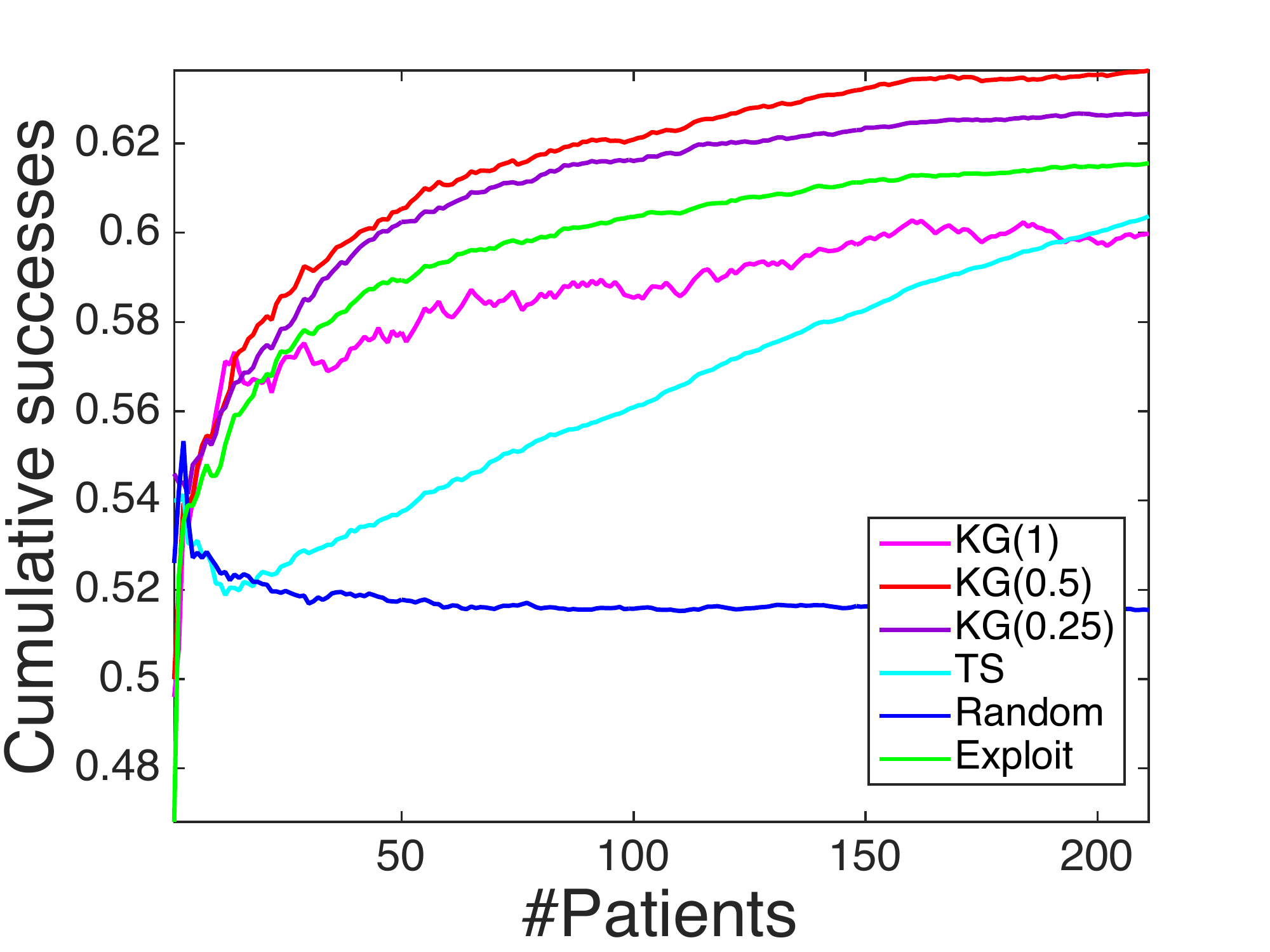}} 
  \subfigure[Distribution of successes. \label{diet}]{   \includegraphics[width=0.45\textwidth]{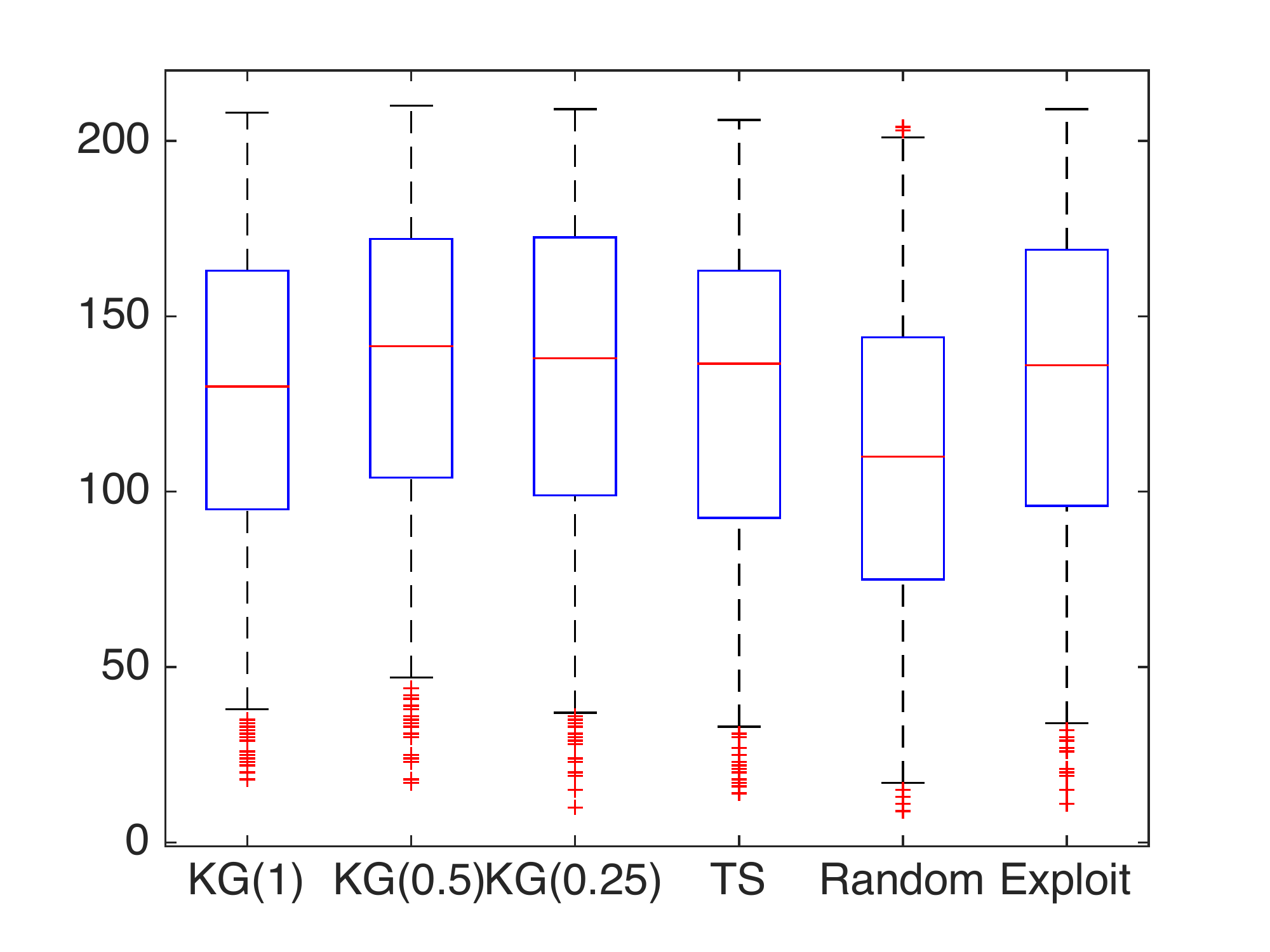}}
\caption{Comparison of different algorithms on the knee replacement dataset. }
\end{figure}
We report the number of cumulative success divided by the number of treated patients after each of the 212 patient visits in Fig. \ref{patient}. We also report the distribution of the number of successes produced by each policy after the learning budget is exhausted in Fig. \ref{diet}.  On each
box, the central red line is the median, the edges of the box
are the 25th and 75th percentiles, and outliers are plotted
individually.

We can see from the figure that the KG policy with $\eta = 0.5$ yields the best performance. Pure exploitation also does well. This seems at first a bit odd given that the system has no prior knowledge about the true parameter value $\bm{w}^*$. A possible explanation is that the change in context induces some level of exploration.  In terms of posterior reshaping, value of smaller $\eta$ in general yields larger number of successes since it is in favor of exploitation
over exploration. But the price to pay is slightly higher variance as seen in Fig. \ref{diet}.  Random assignment does not perform well since it is not learning from its past experiences while other policies use their past observations to guide the next assignment. We conclude that even though the data is sparse (no two patients are alike), through careful selection of physicians we can improve success rates by around 25 percent.

The real value of the knowledge gradient is its rapid learning, which is especially important in a health setting since if we can learn faster, we can benefit more patients.  The knowledge gradient correctly captures the full value of information, properly balancing exploitation (doing well now) and exploration (learning to do well in the future).  Thompson sampling captures the exploration-exploitation tradeoff only approximately.  The real appeal of Thompson sampling is the ease of computation which is useful in high-frequency internet applications.  Other optimizing policies such as pure exploitation (a greedy  policy based on the prior) or Bayes greedy (a greedy policy based on the posterior, similar to Thompson sampling) do not accurately capture the value of information which requires capturing the value of reducing the uncertainty in the belief.

\section{Conclusion}
In this paper, we consider the problem of personalized medicine which formalizes clinical decision making as a function that maps individual patient information to a recommended treatment. The learner is rewarded by ``successes'' and ``failures''  which can be predicted through an unknown relationship that depends on the patient information and the selected treatment. Each experiment is expensive, forcing us to
learn the most from each experiment. 
The goal is to treat current patients as effectively as possible and correctly identify the better treatment as quickly as possible. We adopt a Bayesian approach  both to incorporate possible prior information and to  update our treatment regime continuously as information accrues, with the potential to allow smaller yet more informative trials and for patients to receive better treatment. By formulating the problem as contextual bandits, we introduce a
knowledge gradient policy using Bayesian logistic regression, for which  an approximation
method is used to overcome computational challenges in finding
the knowledge gradient.  We  provide a detailed study on the problem of how sequentially assignment of physicians/facilities to individual patients can reduce the health care cost.   We use clustering and LASSO to deal with the intrinsic sparsity in health datasets. We  show  experimentally that even though the problem is sparse,  through careful selection of physicians (versus picking them at random), we can significantly improve the success rates.

\footnotesize{
\bibliography{refer}
\bibliographystyle{wileyj.bst}}

\end{document}